# Decision Making for Symbolic Probability*

**Phan H. Giang and Sathyakama Sandilya**
Computer-Aided Diagnosis & Therapy
SIEMENS Medical Solutions
51 Valley Stream Pkwy
Malvern, PA 19355 USA
{*phan.giang,sathyakama.sandilya*}*@siemens.com*

## Abstract

This paper proposes a decision theory for a symbolic generalization of probability theory (SP). Darwiche and Ginsberg [2, 3] proposed SP to relax the requirement of using numbers for uncertainty while preserving desirable patterns of Bayesian reasoning. SP represents uncertainty by symbolic supports that are ordered partially rather than completely as in the case of standard probability. We show that a preference relation on acts that satisfies a number of intuitive postulates is represented by a utility function whose domain is a set of pairs of supports. We argue that a subjective interpretation is as useful and appropriate for SP as it is for numerical probability. It is useful because the subjective interpretation provides a basis for uncertainty elicitation. It is appropriate because we can provide a decision theory that explains how preference on acts is based on support comparison.

## 1 Introduction

In [2, 3] Darwiche and Ginsberg proposed a theoretical framework of symbolic probability (SP). The main idea behind SP is to replace numbers by symbols to express *supports* for events. It becomes necessary when (1) exact and consistent numerical probability cannot be estimated due to lack of information and (2) principles of insufficient reasoning (e.g., maximum entropy) are not appropriate. Being a symbolic counterpart of numerical probability, SP provides basic operations to reason with supports. Symbolic probability induces a partially order on events. It is remarkable that most of desirable patterns of inference thought to be unique to Bayesian reasoning also hold for SP. Moreover, SP is shown to subsume not only standard probability calculus but also a number of important calculi used in AI e.g., propositional logic, non-monotonic reasoning, fuzzy possibility and objection-based reasoning.

An open problem for SP is the formulation of a decision theory whose role is similar to the von Neumann-Morgenstern linear utility theory that makes probability so useful. The goal of this paper is to address this problem. We show that a preference relation on acts in a world described by SP structure could be modeled by a *binary* utility function if it satisfies a number of postulates. The binary utility has been introduced in [7, 6] and shown to work with non-probabilistic calculi e.g., possibility theory and consonant belief function.

This paper is structured as follows. In the next section, SP is reviewed. In section 3, we present a set of postulates that lead to a representation theorem. Comparison with related works is presented in section 4 that is followed by a concluding remark.

We list here a brief glossary of symbols to facilitate reading this paper. $\Omega$ is the set of possible worlds. Capital letters $A, B, C$ are used for subsets of the possible world. Lower case letters $a, b, c$ denote acts. $X$ is the set of prizes that includes the most preferred ($\sharp$), the neutral ($\natural$) and the least preferred ($\flat$) elements. $\mathcal{S}$ is the support set that includes $\top$ as the top and $\bot$ as the bottom. Elements of $\mathcal{S}$ are denoted by Greek letters $\alpha, \beta, \gamma$. A support function uses symbol $\Phi$. $\geq_{\oplus}$ denotes a partial order on the support set. $\triangleright$ denotes a preference relation on acts.

## 2 Symbolic probability

This section reviews the symbolic probability theory developed in [2, 3]. SP is motivated by the reality that information available to an agent is often not enough to commit her to a precise numerical probability function.

---
*We thank Bharat Rao for all encouragement and support. We are also indebted to UAI referees for many constructive comments and criticism.

A subjectivist probabilist would argue from Savage's position [14] that the probability an agent holds is not necessarily a product of statistical information; it is something that could be deduced from her behavior. In fact, this would be possible if the behavior satisfies a number of properties (Savage's axioms). However, it has been found in numerous studies [5, 10] that human behavior systematically violates some of Savage's postulates. Another motivation of SP is to emulate properties of Bayesian reasoning. The Bayesian approach is so successful in AI because it combines many plausible reasoning patterns with an efficient computational framework [12].

A probability structure is a tuple $< \Pr, \Omega, [0,1], +, \times >$ where $\Omega$ is the set of possible worlds[1], $[0,1]$ is the unit interval and $+, \times$ are the arithmetic sum and multiplication. A probability function Pr maps the power set $2^\Omega$ into the unit interval that satisfies $\Pr(\emptyset) = 0$, $\Pr(\Omega) = 1$ and $\Pr(A \cup B) = \Pr(A) + \Pr(B)$ if $A \cap B = \emptyset$. If $\Pr_A$ is the conditional probability function given $A$ then $\Pr(A \cap B) = \Pr(A) \times \Pr_A(B)$.

SP is structurally similar to numerical probability. A set $\mathcal{S}$ that includes at least two specially designated elements $\top$ and $\bot$ is a called a *support set*. A *state of belief* (or support function) is a function from the set of events to the support set i.e., $\Phi : 2^\Omega \to \mathcal{S}$ that satisfies 4 axioms

(A1) $\Phi(\emptyset) = \bot$

(A2) $\Phi(\Omega) = \top$

(A3) $\Phi(A \cup B)$ is a function of $\Phi(A)$ and $\Phi(B)$ for $A, B \subseteq \Omega$ and $A \cap B = \emptyset$

(A4) If $A \subseteq B \subseteq C \subseteq \Omega$ and $\Phi(A) = \Phi(C)$ then $\Phi(A) = \Phi(B) = \Phi(C)$

A *partial* function $\oplus : \mathcal{S} \times \mathcal{S} \to \mathcal{S}$ defined according to axiom A3 i.e., $\alpha \oplus \beta$ is defined if there are disjoint events $A, B$ such that $\Phi(A) = \alpha$ and $\Phi(B) = \beta$ and for such $\alpha, \beta$ $\alpha \oplus \beta \stackrel{\text{def}}{=} \Phi(A \cup B)$.

Relation $\geq_\oplus$ on $\mathcal{S}$ is defined as $\alpha \geq_\oplus \beta$ if there is a support $\gamma$ such that $\alpha = \beta \oplus \gamma$. Clearly, $\geq_\oplus$ is a partial order.

Suppose $\Phi_A$ is the updated belief caused by the acceptance of $A$. An event is called *accepted* if its support is $\top$. A number of conditions are imposed on this operation. For events $A, B, C, D$

(A5) $\Phi_A(B) = \top$ if $\Phi(B) = \top$

(A6) $\Phi_A(B) = \Phi(B)$ if $\Phi(A) = \top$

---
[1] We can think of $\Omega$ in terms of a set of variables $V_0, V_1, V_2, \ldots V_k$. Each $\omega \in \Omega$ is a tuple $< v_0, v_1, v_2, \ldots v_k >$ where $v_i$ is a value of variable $V_i$.

(A7) $\Phi_{A \cup B}(A) \geq_\oplus \Phi(A)$

(A8) If $\Phi_C(A) = \Phi_{B \cap C}(A)$ then $\Phi_C(B) = \Phi_{A \cap C}(B)$

(A9) For $\Phi, \Phi'$ such that $\Phi(A \cup B) = \Phi'(A \cup B)$ then
$\Phi_{A \cup B}(C) \geq_\oplus \Phi_{A \cup B}(D)$ iff
$\Phi'_{A \cup B}(C) \geq_\oplus \Phi'_{A \cup B}(D)$

(A10) $\Phi_{A \cup B}(A)$ is a function of $\Phi(A)$ and $\Phi(A \cup B)$

An *unscaling* operator $\otimes$ is defined by $\Phi(A \cap B) = \Phi_A(B) \otimes \Phi(A)$. $\Phi(A)$ is interpreted as the support for the event $A$. For SP, unlike standard probability, support of an event is not a function of the support of its complement.

Darwiche and Ginsberg proved that axioms $A1 - A10$ imply SP structure $< \Phi, \Omega, \mathcal{S}, \oplus, \otimes >$ has properties similar to ones of numerical probability. In particular, support sum is commutative ($\alpha \oplus \beta = \beta \oplus \alpha$), associative ($\alpha \oplus (\beta \oplus \gamma) = (\alpha \oplus \beta) \oplus \gamma$) and absorptive (if $\alpha \oplus \beta \oplus \gamma = \alpha$ then $\alpha \oplus \beta = \alpha$). $\Phi(\emptyset) = \bot$ and $\Phi(\Omega) = \top$. Support unscaling is commutative and distributive, $\alpha \otimes \beta = \beta \otimes \alpha$, $(\alpha \otimes \beta) \otimes \gamma = \alpha \otimes (\beta \otimes \gamma)$. $\bot \otimes \alpha = \bot$, $\alpha \otimes \top = \alpha$.

They have shown that with appropriately defined support domain $\mathcal{S}$, operations $\oplus, \oslash$ and $\otimes$, SP turns into calculi extensively used in AI: probability, propositional logic, fuzzy possibility/Spohn's disbelief calculus, non-monotonic logic based on preferential models and objection-based reasoning.

We conclude this review section with an example of using SP to represent the Ellsberg paradox [5]. This example serves a dual purpose. On one hand, it illustrates the use of SP for situations that elude standard probability. On the other hand, it demonstrates that SP has a subjective interpretation in the same way as standard probability.

The experiment that Ellsberg has set up is simple. In an urn there are 90 balls of three colors Red, White and Black. It is known that 30 balls are red. However nothing is known about the composition of whites and blacks except that together they count 60. Bets are offered to estimate a subject's personal probabilities of White and Black. It is found that the subject strictly prefers a bet on Red (pays $1 if a randomly selected ball is Red and 0 otherwise) to a bet on White. She also strictly prefers the bet on Red to a bet on Black. At the same time, the subject strictly prefers a bet on White or Black (pays $1 if the ball is either White or Black, it pays 0 in case the ball is Red) to a bet on Red or White (pays $1 if the ball is either Red or White and 0 if it is Black). She also prefers the bet on White or Black to a bet on Red or Black.

Clearly, no numerical probability (for White and Black) could be inferred from subject's behavior. Indeed, from the preference of the bet on Red to the bet on White one should conclude that $\frac{1}{3} = P(\text{Red}) > P(\text{White})$. From the preference of the bet on Red to the bet on Black, one must conclude that $\frac{1}{3} = P(\text{Red}) > P(\text{Black})$. These conclusions are not reconcilable with the fact that $P(\text{White or Black}) = \frac{2}{3}$. To make the matter more confused, the first preference (the bet on Red to the bet on White) and the fourth preference (bet on White or Black to the bet on Red and Black) violate Savage's independence axiom.

| Event | ∅ | R | W | B | RW | RB | WB | Ω |
|---|---|---|---|---|---|---|---|---|
| Φ | ⊥ | 1/3 | $w$ | $b$ | $rw$ | $rb$ | 2/3 | ⊤ |

Table 1: Belief state

We use SP to represent this situation. The set of symbolic supports is $\mathcal{S} = \{\bot, w, b, \frac{1}{3}, rw, rb, \frac{2}{3}, \top\}$. One for each event. Belief state $\Phi$ is defined in table 1, where R stands for event "ball is Red", WB stands for "ball is either White or Black" etc. A partial order $\succeq$ can be extracted from the subject's preference on bets. Obviously, $A \succeq \emptyset$ and $\Omega \succeq A$ for any event $A$. The problem here is how to infer a relation on supports from the preference on bets. For example, we know that bet on Red is preferred to the bet on White. It is also preferred to the bet on Black. Obviously, the set of prizes are the same $\{\$1, \$0\}$. So the argument for preference boils down to comparing supports assigned to events. Notice that this is a comparison of two pairs of supports $\langle \frac{1}{3}, \frac{2}{3} \rangle$ and $\langle w, rb \rangle$. In the context of this example, it is possible to argue that since $\$0$ is to be ignored, behind the decision maker choice is her judgment that support for $R$ is larger than support for $W$. Thus, we have $\Phi(R) \geq_\oplus \Phi(W)$ and $\Phi(R) \geq_\oplus \Phi(B)$[2]. The preference of the bet on White or Black to the bet on Red or Black infers $\Phi(WB) \geq_\oplus \Phi(RB)$. Similarly, we have $\Phi(WB) \geq_\oplus \Phi(RW)$. Here $\geq_\oplus$ is a partial order because, for example, W and B are not comparable, so are RW and RB[3].

---

[2] Strictly speaking, order $\frac{1}{3} \geq_\oplus w$ is arrived in two steps. Given that the agent strictly prefers the bet on Red to the bet on White, one can add a "bonus" to the latter so that the agent is indifferent between the original bet on Red and the bet on White plus the bonus. For example, the new bet pays $\$1$ if the selected ball is White, otherwise it gives a chance to roll a die. If the die roll turns 1 (event $O$) then the agent gets $\$1$ otherwise nothing. We conclude from the indifference $\Phi(R) = \Phi(W \cup O) = \frac{1}{3}$. Since $\Phi(W \cup O) \geq_\oplus \Phi(W)$ by definition, we have $\Phi(R) \geq_\oplus \Phi(W)$. Besides providing a formal explanation of the order, O does not provide any information relevant to estimation of W and B. Therefore, such auxiliary symbols will be left out of analysis.

[3] This partial order reflects currently available information. For example, W and B are not comparable because subject's preference between a bet on White and a bet on Black is not revealed. It is possible and necessary that partial order $\succeq$ is updated when more information arrives.

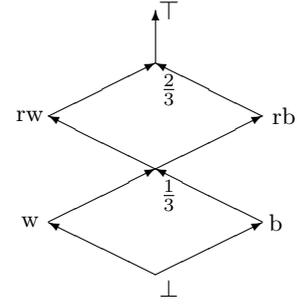

Figure 1: Partial support order

The partial order on belief is supported by the following partial order $\geq_\oplus$ on supports. $\alpha \geq_\oplus \bot$ and $\top \geq_\oplus \alpha$ for any $\alpha \in \mathcal{S}$. $\frac{1}{3} \geq_\oplus w$ and $\frac{1}{3} \geq_\oplus b$. $rw \geq_\oplus \frac{1}{3}$. $rb \geq_\oplus \frac{1}{3}$. $\frac{2}{3} \geq_\oplus rw$ and $\frac{2}{3} \geq_\oplus rb$.

## 3 Decision making with symbolic probability

The previous example with the Ellsberg paradox shows that standard probability does not have monopoly on subjectivistic interpretation. It is important for SP because the subjective interpretation opens the possibility of gaining knowledge about uncertainty from behavior observation. In this section, we ask a different question. Given a description of the world in terms of SP, can one intelligently make decisions on its basis?

We follow the approach in [7, 8]. The basic idea is to reduce acts into equivalent acts of simpler structure that are amenable to the Pareto comparison. Specifically, we assume a SP structure $< \Phi, \Omega, \mathcal{S}, \oplus, \otimes >$. Following Schmeidler [15] we assume that $\Omega$ is constructed from two independent variables $\mathcal{H}$ (for *horse*) and $\mathcal{R}$ (for *roulette*). However, unlike Schmeidler, we do not require that belief state restricted on $\mathcal{R}$ is standard probability. An $\mathcal{H}$−event ($\mathcal{R}$−event) is a subset of $\Omega$ that is expressible by $\mathcal{H}$ ($\mathcal{R}$) only. The independence between $\mathcal{H}$ and $\mathcal{R}$ is satisfied by condition $\Phi_A(B) = \Phi(B)$ and $\Phi_B(A) = \Phi(A)$ where $A$ is a $\mathcal{H}$−event and $B$ is a $\mathcal{R}$−event. In the previous example, $\mathcal{H}$ is *Ball* and $\mathcal{R}$ is *Die*.

We are interested in preference relation on acts. The set of acts $\mathcal{A}$ is constructed as follows. We start with a set of prizes $X$ that is finite and includes at least 3 elements: the best element ♯, the neutral ♮ and the worst element ♭. Intuitively, these elements are representative of 3 categories of prizes: gain, neutral and loss. Each $x \in X$ is a constant act. An one-stage act is formed by assigning a prize $x_i$ for each event $A_i$ in a partition $(A_i)$ of $\Omega$. A two-stage act is formed by assigning either a constant act (prize) or an one-stage act for each event in some partition. If $a_1, a_2, \ldots a_k$ are constant or one-stage acts and $(A_i)_{i=1}^{i=k}$ is a partition of $\Omega$ then tuple $\langle A_i{:}a_i\rangle_{i=1}^{k} \in \mathcal{A}$. An act is interpreted as a contract that rewards the decision maker with $a_i$ when event $A_i$ occurs.

After observation of a non-empty event $A$, an act is transformed into an $A$−act. The conditionalization $(|)$ of acts is defined recursively as follows. The conditionalization of a prize $x \in X$ is itself $(x|A \stackrel{\text{def}}{=} x)$. For any other act: $\langle A_i{:}a_i\rangle_{i=1}^{k}|A \stackrel{\text{def}}{=} \langle (A \cap A_i){:}(a_i|A)\rangle_{i=1}^{k}$ (conditionalization causes readjustment in both events and rewards of the original act).

Two subsets of $\mathcal{A}$ are of special interest. An act where the events on which rewards are conditioned are $\mathcal{H}$−events is called $\mathcal{H}$−act. In other words, $x \in X$ is $\mathcal{H}$−act. If $a_i$ are $\mathcal{H}$−acts and $A_i$ are $\mathcal{H}$−events for $1 \leq i \leq k$ then $\langle A_i{:}a_i\rangle_{i=1}^{k}$ is a $\mathcal{H}$−act. An act is *canonical* if it has the form $\langle G : \sharp, N : \natural, L : \flat \rangle$ where $(G, N, L)$ is a partition of $\Omega$. $G$ is called the *best*, $N$ - the *neutral* and $L$ - the *worst* events of the act.

We consider a collection of preference relations $\{\triangleright_A$ such that $A$ is an $\mathcal{H}$-event$\}$ (reads "at least as preferred as given A"). When we talk about a generic relation in the collection, the subscript is omitted. In terms of this relation, the meaning of $\sharp$ and $\flat$ is that for all $x \in X$, $\sharp \triangleright x$ and $x \triangleright \flat$. What properties do we want for the preference relation?

The first property we desire for $\triangleright$ is transitivity i.e., for $a_1, a_2, a_3 \in \mathcal{A}$ if $a_1 \triangleright a_2$ and $a_2 \triangleright a_3$ then $a_1 \triangleright a_3$. The transitivity ensures basic consistency in subject's behavior. We use symbol $a \bowtie a'$ (reads equivalent) for $a \triangleright a'$ and $a' \triangleright a$.

Another property often imposed on a preference relation is completeness i.e., for any acts $a_1, a_2 \in \mathcal{A}$ decision maker can decide whether $a_1 \triangleright a_2$ or $a_2 \triangleright a_1$. For example, in linear utility theory, the comparison between two acts reduces to comparing their expected utilities (real numbers). However, here we are dealing with the situations when one is not always able to compare supports of events. Is it reasonable to require one to be resolute in choosing acts despite having vague beliefs? We argue that such a requirement would go against the basic tenet of the probabilistic subjectivism: belief in events is inferred from behavior. For example, let us consider two incomparable events $A$ and $B$ i.e., $\Phi(A) \not\geq_\oplus \Phi(B)$ and $\Phi(B) \not\geq_\oplus \Phi(A)$. We consider two acts. One pays \$1 if $A$ occurs and 0 otherwise. The other pays \$1 if $B$ and 0 otherwise. Because we can see no rationale to *justify* definite choice between those bets, therefore, the most natural course would be to let them be incomparable. It is crucial to retain the ability to infer order on supports (e.g., $\Phi(A) \geq_\oplus \Phi(B)$) from preference on bets (e.g., bet on $A$ over the bet on $B$). In sum, we will *not* require the completeness for $\triangleright$ on $\mathcal{A}$.

Between preference relations we assume a condition that imposes consistency. Suppose $a_i \bowtie_{A_i} b_i$ for $i = 1, 2, \ldots m$ then $\langle A_i{:}a_i\rangle \bowtie_\Omega \langle A_i{:}b_i\rangle$. Basically, if $a_i$ is equivalent to $b_i$ given $A_i$, then two acts are substitutable under the assumption that $A_i$ occurs.

Next we consider a requirement that establishes equivalence ($\bowtie$) between different acts. Let us consider, as an example, a two-stage act $a_1 = \langle A{:}a_{11}, \overline{A}{:}a_{12}\rangle$. It means if $A$ occurs one gets act $a_{11}$. If $A$ does not occur then one gets act $a_{12}$. Act $a_{11} = \langle B{:}x_1, \overline{B}{:}x_2\rangle$ delivers $x_1$ if $B$ occurs and $x_2$ if $B$ does not. Act $a_{12} = \langle C{:}x_1, \overline{C}{:}x_2\rangle$ delivers $x_1$ if $C$ occurs and $x_2$ if $C$ does not. By collapsing two stages into one, we have a new act $a_2 = \langle D{:}x_1, \overline{D}{:}x_2\rangle$ where $D = (A \cap B) \cup (\overline{A} \cap C)$. $a_2$ delivers $x_1$ if $D$ and $x_2$ if $\overline{D}$. We require that $a_1 \bowtie_\Omega a_2$.

In order to relate prizes to supports, we require that given the acceptance of an $\mathcal{H}$−event $A$, each prize $x \in X$ is equivalent to a $\mathcal{R}$−canonical act i.e., $x \bowtie_A \langle G_x : \sharp, N_x : \natural, L_x : \flat \rangle$ where $G_x, N_x, L_x$ are $\mathcal{R}$−events. Notice that because of independence between $\mathcal{H}$ and $\mathcal{R}$, the triplet does not depends on $A$. By the same reason, conditional supports $\Phi_A(G_x)$, $\Phi_A(N_x)$ and $\Phi_A(L_x)$ are invariant wrt $A$.

Finally, we postulate a fundamental connection between support and choice: comparing two canonical acts, decision maker (1) prefers one whose best event is more supported and whose worst event is less supported; (2) ignores the neutral events of both acts. Formally, $\langle A_1{:}\sharp, A_2{:}\natural, A_2{:}\flat\rangle \triangleright_A \langle B_1{:}\sharp, B_2{:}\natural, B_2{:}\flat\rangle$ iff $\Phi(A_1) \geq_\oplus \Phi(B_1)$ and $\Phi(B_3) \geq_\oplus \Phi(A_3)$. This Paretian argument formalizes intuitive notion that people desire gain, avoid loss and do not care about things that have no effect on their utility. The linear utility theory exhibits an analogous behavior. The partial derivative of expected utility on event probability is the utility of reward obtained in case of the event. The probability assigned to the event leading to zero utility does not contribute to the expected utility. Probabilities of events leading to positive (negative) utility positively (negatively) affect the expected utility.

Formally, we assume that $\{\triangleright_A\}$ satisfy the following postulates that are similar to ones used by Luce and Raiffa [11] for linear utility theory.

D1 $\triangleright_A$ is a partial order on $\mathcal{A}$ for any $A \subseteq \Omega$.

D2 Suppose $a_i = \langle B_{ij}{:}x_j\rangle_{j=1}^m$ for $1 \leq i \leq n$, $(A_i)_{i=1}^n$ is a partition of $A$ then $\langle A_i{:}a_i\rangle_{i=1}^n \bowtie_A \langle C_j{:}x_j\rangle_{j=1}^m$ where
$$C_j = \cup_{i=1}^n (A_i \cap B_{ij}) \qquad (1)$$

D3 For each $x \in X$ and a non-empty $\mathcal{H}$−event

$A$ there exists a $\mathcal{R}$−triplet partition of $\Omega$: $(G_x, N_x, L_x)$ such that $x \bowtie_A \langle G_x{:}\sharp, N_x{:}\natural, L_x{:}\flat\rangle$.

D4 If $(A_i)_{i=1}^n$ is a partition of $A$ and $a_i' \bowtie_{A_i} a_i$ for $1 \leq i \leq n$ then $\langle A_i{:}a_i\rangle_{i=1}^n \bowtie_A \langle A_i{:}a_i'\rangle_{i=1}^n$.

D5 $\langle G : \sharp, N : \natural, L : \flat\rangle \;\triangleright_C\; \langle G' : \sharp, N' : \natural, L' : \flat\rangle$ iff $\Phi_C(G) \geq_\oplus \Phi_C(G')$ and $\Phi_C(L') \geq_\oplus \Phi_C(L)$

A desirable property for a preference relation is the *uniformity* with respect to support values [9]. Relation $\triangleright_A$ is *uniform* if it satisfies the following condition. For partitions $(A_i)_{i=1}^m$ and $(B_i)_{i=1}^m$ of $A$ such that $\Phi_A(A_i) = \Phi_A(B_i)$, $\langle A_i{:}x_i\rangle_{i=1}^m \bowtie_A \langle B_i{:}x_i\rangle_{i=1}^m$ for $x_i \in X$. This condition simply says that the compositions of events do not matter. What does matter is the supports attached to those events. An implication is that we can abstract away the events of an act and replace them with support values. We use notation $[\phi_i{:}x_i]_{i=1}^m$ for the class of acts $\langle A_i{:}x_i\rangle_{i=1}^m$ such that $\Phi(A_i) = \phi_i$ for $i = 1, 2, \ldots m$. Such a class of acts is called a *lottery*. Postulate D5 allows us to simplify further notation for canonical lotteries. Since the supports of the neutral events of canonical lotteries do not affect the preference they can be left out. $[\Phi(G){:}\sharp, \Phi(N){:}\natural, \Phi(L){:}\flat]$ is abbreviated by a pair of supports $\langle \Phi(G), \Phi(L)\rangle$.

**Theorem 1** *Let $C$ be a $\mathcal{H}$−event and $(A_i)_{i=1}^m$ and $(B_i)_{i=1}^m$ be two $\mathcal{H}$−partitions of $C$ such that $\Phi_C(A_i) = \Phi_C(B_i)$. If collection $\{\triangleright_A\}$ satisfies postulates $D1-D5$ then $\langle A_i{:}x_i\rangle_{i=1}^m \bowtie_C \langle B_i{:}x_i\rangle_{i=1}^m$.*

**Proof:** We name the acts in question by $d_A$ and $d_B$ respectively. By $D3$, we have $x_i \bowtie_{A_i} a_i$ and $x_i \bowtie_{B_i} a_i$ where $a_i = \langle G_i : \sharp, N_i : \natural, L_i : \flat\rangle$ for $i = 1, 2, \ldots m$. Also by $D3$ we have $\Phi_{A_i}(G_i) = \Phi_{B_i}(G_i)$, $\Phi_{A_i}(N_i) = \Phi_{B_i}(N_i)$ and $\Phi_{A_i}(L_i) = \Phi_{B_i}(L_i)$. By $D4$, we can substitute $a_i$ for $x_i$ in $d_A$ and $d_B$ to get $d_A'$ and $d_B'$. By $D2$, we can collapse $d_A'$ and $d_B'$ to get $d_A''$ and $d_B''$. $d_A''$ is a canonical act $\langle G_A : \sharp, N_A : \natural, L_A : \flat\rangle$ and $d_B''$ is a canonical act $\langle G_B : \sharp, N_B : \natural, L_B : \flat\rangle$ where

$$G_A = \cup_{i=1}^m (A_i \cap G_i), \; G_B = \cup_{i=1}^m (B_i \cap G_i)$$
$$L_A = \cup_{i=1}^m (A_i \cap L_i), \; L_B = \cup_{i=1}^m (B_i \cap L_i)$$

We calculate the supports for $G_A$ and $G_B$:

$$\Phi_C(G_A) = \oplus_{i=1}^m (\Phi_C(A_i) \otimes \Phi_{A_i}(G_i))$$
$$\Phi_C(G_B) = \oplus_{i=1}^m (\Phi_C(B_i) \otimes \Phi_{B_i}(G_i))$$

We have $\Phi_C(A_i) = \Phi_C(B_i)$ by the theorem conditions and $\Phi_{A_i}(G_i) = \Phi_{B_i}(G_i)$ by postulate $D3$. Thus, $\Phi_C(G_A) = \Phi_C(G_B)$. Similarly, we can show that $\Phi_C(L_A) = \Phi_C(L_B)$. By $D5$, we have $d_A'' \bowtie_C d_B''$. Since $d_A \bowtie_C d_A' \bowtie_C d_A''$ and $d_B \bowtie_C d_B' \bowtie_C d_B''$, by $D1$ we conclude $d_A \bowtie_C d_B$. ∎

**Theorem 2** *Suppose $\langle A_i{:}a_i\rangle_{i=1}^m$ is a $\mathcal{H}$−act (i.e., $(A_i)$ is a $\mathcal{H}$−partition of $A$ and $a_i$ are $\mathcal{H}$−acts), then there is a triplet partition $(G, N, L)$ such that $\langle A_i{:}a_i\rangle_{i=1}^m \bowtie_A \langle G : \sharp, N : \natural, L : \flat\rangle$.*

**Proof:** The technique used in the previous proof can also be used here. We sketch a proof by induction on the depth of the act. The induction hypothesis is that there exist canonical acts such that

$$a_i \bowtie_{A_i} \langle G_i : \sharp, N_i : \natural, L_i : \flat\rangle \text{ for } i = 1, 2, \ldots m$$

If $a_i$ are prizes the hypothesis holds because of $D3$. Assume that it holds for acts of depth less than $n$ where $n$ is the depth of the act in question. We substitute the canonical acts for $a_i$ by $D4$, then collapse the newly formed act by $D2$ to obtain $\langle G : \sharp, N : \natural, L : \flat\rangle$ where

$$G = \cup_{i=1}^m (A_i \cap G_i) \qquad (2)$$
$$N = \cup_{i=1}^m (A_i \cap N_i) \qquad (3)$$
$$L = \cup_{i=1}^m (A_i \cap L_i) \qquad (4)$$

The transitivity $(D1)$ of $\triangleright$ will conclude the proof. ∎

This theorem clearly outlines a strategy for deciding between two acts: (1) reduce an act into canonical form; (2) compare the equivalent canonical acts on the basis of $D5$. We need some notation. By $D3$, for $x \in X$, $x \bowtie_A \langle G_x : \sharp, N_x : \natural, L_x : \flat\rangle$. Let us define two functions $\alpha, \beta$ from $X$ to $\mathcal{S}$ by

$$\alpha(x) \mapsto \Phi_A(G_x) \text{ and } \beta(x) \mapsto \Phi_A(L_x) \qquad (5)$$

In particular $\alpha(\sharp) = \top$, $\beta(\sharp) = \bot$, $\alpha(\natural) = \bot$, $\beta(\natural) = \bot$, $\alpha(\flat) = \bot$, $\beta(\flat) = \top$ because $G_\sharp = N_\natural = L_\flat = \Omega$ and $N_\sharp = L_\sharp = G_\natural = L_\natural = G_\flat = N_\flat = \emptyset$.

We overload symbols $\oplus$ and $\otimes$ as follows. Suppose $\alpha, \beta, \gamma, \beta', \gamma' \in \mathcal{S}$

$$\alpha \otimes \langle \beta, \gamma\rangle \stackrel{\text{def}}{=} \langle \alpha \otimes \beta, \alpha \otimes \gamma\rangle$$
$$\langle \beta, \gamma\rangle \oplus \langle \beta', \gamma'\rangle \stackrel{\text{def}}{=} \langle \beta \oplus \beta', \gamma \oplus \gamma'\rangle$$

Operations on the left hand side are defined from the support sum and unscaling operations on the right hand side. Thus, they are defined whenever the corresponding support operations are defined. We have the following corollary.

**Corollary 1** *Suppose $(A_i)_{i=1}^m$ is a $\mathcal{H}$−partition of $A$ and $\Phi_A(A_i) = \phi_i$*

$$[\phi_i{:}x_i]_{i=i}^m \bowtie_A \oplus_{i=1}^m (\phi_i \otimes \langle \alpha(x_i), \beta(x_i)\rangle) \qquad (6)$$

Note that the calculation on the right hand side of eq. 6 is well defined because this process is nothing but

calculating the supports for events of the canonical act guaranteed by theorem 2. We can define a set

$$\mathcal{U} \stackrel{\text{def}}{=} \{\langle \alpha, \beta \rangle \,|\, \exists A, B, A \cap B = \emptyset, \Phi(A) = \alpha, \Phi(B) = \beta\}$$

with order $\geq_\mathcal{U}$ defined as $\langle \alpha, \beta \rangle \geq_\mathcal{U} \langle \alpha', \beta' \rangle$ iff $\alpha \geq_\oplus \alpha'$ and $\beta' \geq_\oplus \beta$. A function $u : \mathcal{A} \to \mathcal{U}$ is defined by

$$u(\langle A_i{:}x_i \rangle_{i=i}^m) \mapsto \oplus_{i=1}^m (\Phi_A(A_i) \otimes \langle \alpha(x_i), \beta(x_i) \rangle) \quad (7)$$

$u$ can be seen as a utility function where the utility range is $\mathcal{U}$. Since each utility is a pair of two components, we call the utility *binary*. We refer to the left (right) component of a binary utility as its *gain* (*loss*) component.

It is not difficult to verify that for canonical acts $u(\langle G:\sharp, N:\natural, L:\flat \rangle) = \langle \Phi(G), \Phi(L) \rangle$. Because of eq. 5 for prize $x$, $u(x) = \langle \alpha(x), \beta(x) \rangle$. In particular, $u(\sharp) = \langle \top, \bot \rangle$, $u(\natural) = \langle \bot, \bot \rangle$ and $u(\flat) = \langle \bot, \top \rangle$. Given that eq. 7 becomes

$$u(\langle A_i{:}x_i \rangle_{i=i}^m) \mapsto \oplus_{i=1}^m (\Phi_A(A_i) \otimes u(x_i)) \quad (8)$$

We conclude this section with an illustration of SP solution for the Ellsberg paradox. The acts considered in the example involves a minimal set of prizes. $X$ consists of just 3 elements $\sharp$, $\natural$ and $\flat$. $\text{Bet}_{red}$ that pays \$1 if the ball is Red and nothing otherwise is coded as $\langle R:\sharp, BW:\natural, \emptyset:\flat \rangle$. $\text{Bet}_{white}$ is coded as $\langle W:\sharp, RB:\natural, \emptyset:\flat \rangle$.

$$u(\text{Bet}_{red}) = \begin{cases} \Phi(R) \otimes \langle \top, \bot \rangle \oplus \\ \Phi(WB) \otimes \langle \bot, \bot \rangle \oplus \\ \Phi(\emptyset) \otimes \langle \bot, \top \rangle \end{cases}$$
$$= \begin{cases} \frac{1}{3} \otimes \langle \top, \bot \rangle \oplus \\ \frac{2}{3} \otimes \langle \bot, \bot \rangle \oplus = \langle \frac{1}{3}, \bot \rangle \\ \bot \otimes \langle \bot, \top \rangle \end{cases}$$

Similarly, we calculate $u(\text{Bet}_{white}) = \langle w, \bot \rangle$. Since $\frac{1}{3} \geq_\oplus w$, we have $u(\text{Bet}_{red}) \geq_\mathcal{U} u(\text{Bet}_{white})$.

## 4 Related works

Qualitative decision making has been getting attention of researchers in recent years out of concern that reliable probability may not practically available. Dubois and Prade [4] studied a framework for decision making based on possibility theory. Brafman and Tennenholtz [1] provided an axiomatic characterization of frequently used qualitative decision criteria namely maximin, minimax regret and competitive ratio.

This work continues the development of our previous works on decision making with various forms of non-probabilistic uncertainty calculi [6, 8]. The basic idea is to reduce acts of complex structure into ones of simpler structure (canonical) that are amenable to the Pareto comparison. This approach exploits the duality between belief and preference. A comparison with Dubois and Prade's approach [4] is given in [7]. While qualitative decision criteria such as maximin, minimax regret and competitive ratio ignore uncertainty by focusing on worst-case performance, our approach make use of this information in making decision.

It can be shown that when SP becomes possibility calculus ($\mathcal{S}$ is the real unit interval, $\oplus$ is max and $\otimes$ is min) this proposal is more flexible than one in [7]. In the latter preference relation was required to be complete whereas in this proposal it could be partial. However, if we require $N_x = \emptyset$ for $x \in X$, the utility function defined in eq. 8 becomes the binary utility function for possibility [7].

When SP becomes standard probability ($\mathcal{S}$ is $[0,1]$, $X$ is $[0,1]$, $\oplus$ is $+$, and $\otimes$ is $\times$) we can show that utility function defined in eq. 7 is equivalent to standard expected utility. We set $u(x) = \langle x, 1-x \rangle$ (this also means $N_x = \emptyset$). Therefore, $\langle \alpha, 1-\alpha \rangle \geq_\mathcal{U} \langle \alpha', 1-\alpha' \rangle$ iff $\alpha \geq \alpha'$. Eq. 8 becomes

$$\begin{aligned} u(\langle A_i{:}x_i \rangle_{i=i}^m) &= \sum_{i=1}^m (p(A_i) \times \langle x_i, 1-x_i \rangle) \\ &= \left\langle \sum_{i=1}^m p(A_i)x_i, \sum_{i=1}^m p(A_i)(1-x_i) \right\rangle \\ &= \left\langle \sum_{i=1}^m p(A_i)x_i, (1 - \sum_{i=1}^m p(A_i)x_i) \right\rangle \end{aligned}$$

The gain component of binary utility is the same as standard expected utility. Because, in this case, gain and loss components are complementary, order on binary utilities is the same as order on their gain components. Therefore, the binary utility function is equivalent to the expected utility function.

Partial orders of belief and the issues of decision making have been considered by a number of authors. We examine in this section two of the more prominent proposals. Halpern [9] considers *plausibility* measures[4]. Plausibility function $Pl$ is a function from $2^\Omega \to D$ where $D$ is the set of plausibility values. $D$ is partially ordered by $\geq_D$ (transitive, reflexive and antisymmetric). $\top$ and $\bot$ are top and bottom elements of $D$. $Pl$ is assumed to satisfy 3 properties $Pl(\emptyset) = \bot$, $Pl(\Omega) = \top$ and $Pl(A) \geq_D Pl(B)$ if $A \supseteq B$. He shows that a set of standard probability functions $\mathcal{P}$ can be formulated as a plausibility measure by taking plausibility range $D_\mathcal{P} \stackrel{\text{def}}{=} \{f : \mathcal{P} \to [0,1]\}$ - the set of function from $\mathcal{P}$ to unit interval. Partial order on $D_\mathcal{P}$ is defined as pointwise ordering i.e., $f \geq g$ iff $f(P) \geq g(P) \forall P \in \mathcal{P}$. Plausibility measure is defined as $Pl(A) \mapsto f_A$ where $f_A$ is defined as $f_A(P) = P(A)$ for $P \in \mathcal{P}$. For example, the Ellsberg example is represented by a set of probability functions $\{P_a | 0 \leq a \leq \frac{2}{3}\}$ with the understanding that $P_a(W) = a$. Thus, $Pl(R) = f_{red}$ where $f_{red}(P_a) = \frac{1}{3}$, $Pl(W) = f_{white}$ where $f_{white}(P_a) = a$ for $0 \leq a \leq \frac{2}{3}$ and $Pl(B) = f_{black}$ where $f_{black}(P_a) = \frac{2}{3} - a$. This representation can be viewed as a SP structure with (1) set of support $\mathcal{S} = D_\mathcal{P}$, (2) $\geq_\oplus \ = \geq_D$, (3) operations $\oplus, \otimes, \oslash$ are pointwise addition, multiplication and division respectively. Notice that in Halpern's representation, unlike ours, $f_{red} \not\geq_D f_{white}$ and $f_{red} \not\geq_D f_{black}$.

---

[4]Halpern's notion of plausibility measure is more general than Dempster-Shafer plausibility functions.

To make decision with plausibility, Halpern defines an *expectation domain* which is a tuple $(D_1, D_2, D_3, \oplus, \otimes)$ where $D_1$ is the plausibility range, $D_2$ is the prize value range and $D_3$ is the valuation domain. $\otimes : D_1 \times D_2 \to D_3$ and $\oplus : D_3 \times D_3 \to D_3$. $d_1 \otimes d_2$ is the valuation of having a value $d_2$ with plausibility $d_1$. $d_3 \oplus d_3'$ is the valuation of owning two valuations $d_3$ and $d_3'$. In particular, $D_1 = D_\mathcal{P}$, $D_2$ is the set of *constant* functions from $\mathcal{P}$ to $R$ - the set of reals. $D_3$ is the set of functions from $\mathcal{P}$ to $R$. $\oplus$ is pointwise addition and $\otimes$ is pointwise multiplication. The expectation of an act is defined as

$$E_{Pl}(\langle A_i : x_i \rangle_{i=1}^m) \stackrel{\text{def}}{=} \oplus_{i=1}^m (Pl(A_i) \otimes x_i) \quad (9)$$

The similarity between the expectation expression (eq. 9) and binary utility (eq. 8) is obvious. The expectation maps acts into valuation domain $D_3$. So the comparison of acts reduces to comparison of their valuations. Since valuation domain $D_3$ has no obvious order (in particular, it is a set of functions $D_\mathcal{P}$), a decision maker must provide an order through a mechanism called a *decision rule*. In the case of $\mathcal{P}$, there are several options for the decision rule: (1) $d_3 \geq_1 d_3'$ iff $\inf(d_3) \geq \inf(d_3')$; (2) $d_3 \geq_2 d_3'$ iff $\sup(d_3) \geq \sup(d_3')$; (3) $d_3 \geq_3 d_3'$ iff $\inf(d_3) \geq \sup(d_3')$; and (4) $d_3 \geq_4 d_3'$ iff $d_3(P) \geq d_3'(P))$ for all $P \in \mathcal{P}$. Notice that among 4 rules, only $\geq_1$ is compatible with the preference observed in the Ellsberg example i.e., $\langle R:\sharp, BW:\natural \rangle \geq_1 \langle W:\sharp, RB:\natural \rangle$ and $\langle R:\sharp, BW:\natural \rangle \geq_1 \langle B:\sharp, RW:\natural \rangle$. But this rule also forces some preferences which are not observed, for example, $\langle RW:\sharp, B:\natural \rangle =_1 \langle R:\sharp, WB:\natural \rangle$.

Although our postulate D5 could be viewed as a decision rule. Its justification would be impossible without interpretation based on the supports of the best and the worst events of canonical acts. In Halpern's approach, the flexibility offered by including decision rules and expectation domain into decision making process is gained at the cost of divorcing belief structure from preference observation. For example, $Pl(R)$ and $Pl(W)$ are not comparable while the bet on Red is definitely preferred to the bet on White. Operations $\oplus, \otimes$ of the expectation domain are independent from the operations on belief. In contrast, our approach is based on the duality of belief and choice preference where the operations for utility are the same as the operations on belief.

Schmeidler [15] and Sarin & Wakker [13] developed a decision theory on the basis of Choquet expected utility. Their works extend Savage's approach [14]. They show that the satisfaction of a number of postulates by preference on acts leads to the existence of a non-additive (subjective) measure of uncertainty called *capacity*. Essentially, a capacity function is a Halpern plausibility measure whose range is the unit interval. Schmeidler [15] considers a set-up in which acts have two stages. In the first stage, uncertainties of events are capacity and rewards are probabilistic acts whose events have standard (additive) probabilities. Sarin and Wakker [13] consider one-stage acts. The set of events on which the acts are formed is required to contain a subset of *unambiguous* events. On this subset, capacity is standard probability.

Let us assume that prizes are real numbers interpreted as utility. Prize order is $x_1 \geq x_2 \geq \ldots \geq x_m$. Uncertainty is represented by capacity function $v$. The Choquet expected utility (CEU) of acts is defined as

$$CEU(\langle A_i : x_i \rangle_{i=1}^m) = \sum_{i=1}^m x_i(v_i - v_{i-1}) \quad (10)$$

where $v_0 = v(\emptyset) = 0$ and $v_i = v(A_1 \cup A_2 \ldots \cup A_i)$.

This approach upholds the principle of subjectivism that allows inferring belief from the subject's preference. For example, preference of the bet on Red to the bet on White implies that the capacity of White is a number less than $\frac{1}{3}$. To account for violations of Savage's axioms, capacity measure is allowed to be non-additive i.e., for $A \cap B = \emptyset$, $v(A) + v(B) \neq v(A \cup B)$. In particular, $v(A) + v(\overline{A}) \neq 1$. Since addition (+) does not apply for capacity, the only benefit (also the goal) of using reals is to retain multiplication ($\times$) for uncertainty-utility combination.

To see the difference between the binary utility and CEU, let us assume, for the sake of comparison, $\sharp = 1$, $\natural = 0$, $\flat = -1$ and $\mathcal{S} = [0, 1]$. With capacity $v$ CEU of $\langle G:\sharp, N:\natural, L:\flat \rangle$ is $v(G) + v(GN) - 1$. With support function $\Phi$, binary utility of the same act is $\langle \Phi(G), \Phi(L) \rangle$. It can be shown that in a special case when support and capacity are identical ($\Phi = v$) and are additive (standard probability), binary utility and Choquet utility are equivalent. Normally, they will induce different orderings. Note also that capacity of $L$ does not affect CEU (capacity is not additive therefore $v(L) \neq 1 - v(GN)$). This insensitivity to uncertainty of losses could be a practical problem of using CEU.

Since capacity as well as CEU values are reals, orders on belief and on utility are complete. This implies that a subject is not allowed to be indecisive, which is not the case for our approach with SP. Another issue is that Schmeidler's two-stage approach and Sarin and Wakker's one-stage approach are irreconcilable [13]. In our model, two-stage and one-stage views are equivalent due to postulate $D2$. From computational point of view, capacity, as a measure of uncertainty, is probably less useful because it is not clear if and how conditionalization operation could be defined.

## 5 Conclusion

In this paper, we investigate decision making with the symbolic generalization of probability proposed by Darwiche and Ginsberg to relax the commitment to numerical values while preserving desirable properties of the Bayesian approach. If we assume that the preference relation on acts satisfies a number of postulates then it is represented by a utility function whose range is a set of pairs of symbols in the support domain. Two components of binary utility can be thought of as the supports attached to the best and the worst events of a canonical act. This interpretation is useful to justify utility order. The higher the gain component and the lower the loss component the better. The binary utility function has the same structure as the expected utility function. This work provides a subjective interpretation for SP. It does not answer, however, a reverse question what behavior characterizes the SP structure.

We are investigating applications of these techniques to the domain of medical diagnosis where the challenge in exact numeric knowledge elicitation is well known. In such scenarios, we believe that our technique can be used to elicit preferences from physicians based on choices they make regarding diagnostic tests or therapeutic procedures, and use these preferences to construct a consistent decision making framework. Moreover, such a framework may be arrived at using expert physicians at research institutions, and then can be used as a diagnostic aid for other physicians.